\pdfoutput=1

\documentclass[11pt]{article}

\usepackage[final]{acl}

\usepackage{times}
\usepackage{latexsym}
\usepackage{amsmath}
\usepackage{float}
\usepackage{subcaption}

\usepackage{booktabs}

\usepackage[T1]{fontenc}
\usepackage{graphicx}
\usepackage{multirow}
\usepackage{lipsum} 
\usepackage{array}  
\usepackage{subcaption}
\usepackage{xfrac}

\usepackage[utf8]{inputenc}

\usepackage{microtype}

\usepackage{inconsolata}

\title{Quality Matters: Evaluating Synthetic Data for Tool-Using LLMs}

\author{Shadi Iskander\thanks{This work was done during an internship in Amazon and as part of graduate studies at the Technion - Israel Institute of Technology.} \\
  Amazon\\
  \texttt{shadisk@amazon.com} \\
  \And
  Nachshon Cohen \\
  Amazon \\
  \texttt{nachshon@amazon.com} \\
  \And
  Zohar Karnin \\
  Technology Innovation Institute \\
   \texttt{zohar.karnin@tii.ae} \\
   \AND
   Ori Shapira \\
   OriginAI \\
   \texttt{obspp18@gmail.com} \\
   \And
  Sofia Tolmach \\
  Amazon \\
  \texttt{sofiato@amazon.com} \\
  }

\newcommand{\evaluatorModel}{ChatGPT}

\begin{document}

\maketitle
\begin{abstract}
Training large language models (LLMs) for external tool usage is a rapidly expanding field, with recent research focusing on generating synthetic data to address the shortage of available data. However, the absence of systematic data quality checks poses complications for properly training and testing models. To that end, we propose two approaches for assessing the reliability of data for training LLMs to use external tools. The first approach uses intuitive, human-defined correctness criteria. The second approach uses a model-driven assessment with in-context evaluation. We conduct a thorough evaluation of data quality on two popular benchmarks, followed by an extrinsic evaluation that showcases the impact of data quality on model performance. Our results demonstrate that models trained on high-quality data outperform those trained on unvalidated data, even when trained with a smaller quantity of data. These findings empirically support the significance of assessing and ensuring the reliability of training data for tool-using LLMs.

\end{abstract}
\section{Introduction}
\label{sec_introduction}

Enabling LLMs to make use of external tools is a promising frontier that allows tapping into information that is not readily available to the model itself
\cite{huang2023metatool, li-etal-2023-api,qin2024toolllm,tang2023toolalpaca,yang2023auto,patil2023gorilla,schick2024toolformer}. Given a request and a list of available external API functions, the basic task of a model is to collect information by invoking functions, and then to generate a response for the request.
Due to the lack of data for the task and the high cost of creating such data, researchers have
devised synthetic datasets, predominantly with the assistance of LLMs
\cite{huang2023metatool, li-etal-2023-api, tang2023toolalpaca}. These datasets have facilitated a great leap in promoting the appealing applications of tool-using LLMs.

Recently, \citet{zhou2024lima} showed that higher quality training data yields better performance by LLMs in text generation tasks.
However, leading works on tool-using LLMs have not made an effort to measure the quality of training data. Rather, only model outputs are extrinsically evaluated, disregarding the effect of the data on the tested models.
Most research on tool-using LLMs focuses on improving training and evaluation processes \cite{huang2023metatool, qin2024toolllm, tang2023toolalpaca}. The lack of attention to \textit{data quality} makes it difficult to interpret potential pitfalls for models. In turn, this wastes valuable resources for configuring and tuning models over possibly erroneous data.
\begin{figure}[t]
  \centering
  \includegraphics[width=0.8\linewidth]{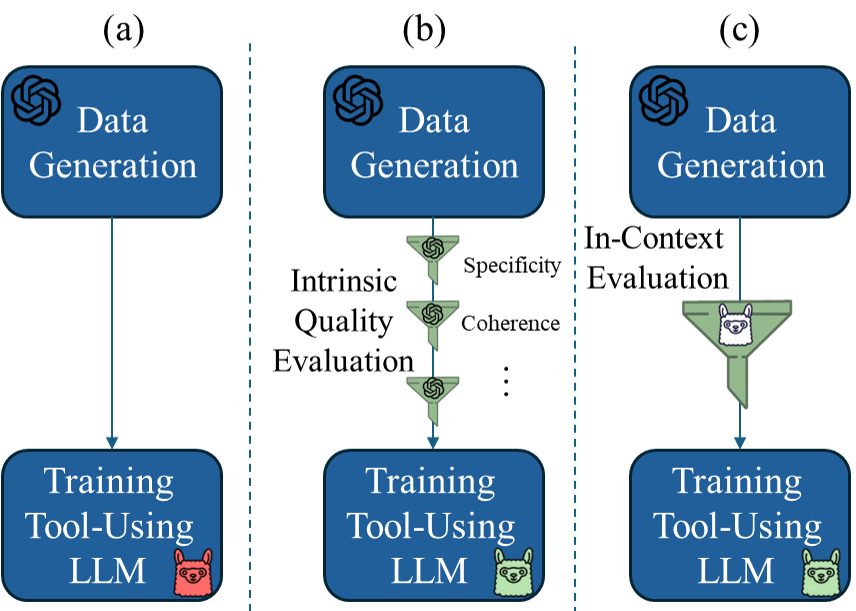}
  \caption{Data quality assessment methods for improving the training 
  process of tool-using LLMs (a), employing two different approaches: (b) intrinsic quality evaluation, using an external LLM to measure various human-defined criteria;
 (c) in-context evaluation, using the target LLM to measure the educational value of data instances. A smaller high-quality training dataset is more effective than a larger unvalidated set.}
  \label{fig:intro}
  \vspace{-7pt}
\end{figure}

Datasets for tool-using LLMs comprise instructions and ground truth API call sequences, and are created mainly with LLMs.
Two such prominent datasets \citep{qin2024toolllm, tang2023toolalpaca} were produced with the help of \evaluatorModel{} \citep{openai2024gpt4}, and were not explicitly assessed for their quality. A closer inspection, conducted in this work, reveals numerous errors within the data, both in the instructions and in the ground-truth API calls (\S\ref{sec_intrinsic}). \looseness=-1

To conduct our inspections, we define intrinsic measures for data quality assessment,
focusing on different aspects of quality. For each aspect we outline human evaluation guidelines,
as well as implement automated methods for evaluation.
The automatic methods employ \evaluatorModel{}, either by directly asking for its evaluation or by having it perform a proxy task and deriving the evaluation from its output. We show high agreement for our automated methods with expert human annotations. 
%
In addition to the intrinsic measures, we propose a metric we call \textit{In-Context Evaluation} (ICE; \S\ref{sec_ice}). ICE evaluates a data instance by how helpful it is for in-context learning, thus predicting its helpfulness for training a model (\S\ref{sec_ice}). 
This metric is fully automated and does not rely on task-specific measurement definitions.

Other than being appraisal instruments, the intrinsic evaluation and ICE metrics can be used to automatically filter out low quality data from an existing dataset. In Section \ref{sec_extrinsic_eval} we carry out this procedure, and display the effect of training tool-using LLMs with higher quality data.
Our findings, demonstrated
on the ToolBench \cite{qin2024toolllm} and ToolAlpaca \cite{tang2023toolalpaca} benchmarks, show either better or comparable performance when using a small high-quality training dataset, compared to the original models trained on larger unverified datasets. The two benchmarks 
are based on different API function sets, and different data generation and training methods, indicating the generalized applicability of our methods.

\section{Background and Related Work}
\label{sec_related_work}

\subsection{External Tool Usage by LLMs}
Tool learning is a recent area of research, aiming to enable LLMs to overcome limitations by accessing tools for, e.g., retrieving up-to-date information \citep{kasai2023realtime, cheng2024dated}, or performing mathematical calculations \citep{schick2024toolformer}, thereby enhancing their usability for real-world needs.\looseness=-1

Research on tool learning focuses on various aspects of training LLMs to use external tools. These mainly include tool selection, tool usage, and planning \citep{zhuang2024toolqa,qin2024toolllm, patil2023gorilla,yao2023react}.
Such models are mainly evaluated extrinsically, only measuring the final results. T-Eval \citep{chen2024teval} is the first evaluation framework that analyzes tool-using LLMs intrinsically.  That is, it decomposes the evaluation into all sub-tasks (such as selection, usage and planning), measuring the fine-grained abilities of models as
tool agents.
We intrinsically evaluate the \textit{data} for tool-usage instead of a \textit{model}.

\subsection{Data Generation}
\label{sec:related_gen}
Recent notable works generated synthetic data for tool learning. 
ToolBench \cite{qin2024toolllm} leverages a large pool of real API functions.\footnote{Based on RapidAPI: \url{https://rapidapi.com/hub}} ChatGPT \citep{openai2024gpt4} was used to generate an instruction that would require invoking a given small set of these tools, as well as to produce a solution path for the respective instruction.
The data was constructed with a varying number of tools per instance and varying relatedness between the tools. API Bank \citep{li-etal-2023-api} created synthetic API documentation, instruction queries, and responses using strong LLMs \citep[GPT-4 and ChatGPT;][]{openai2024gpt4}. A smaller test set was created and validated manually by humans. 
\citet{tang2023toolalpaca} constructed the ToolAlpaca dataset using ChatGPT to generate cleaner documentation upon existing APIs, and respective instructions and responses. In ToolAlpaca, most synthesized instructions only require a single tool to fulfill the request.
The test set was validated by humans to ensure~quality.

To strengthen the credibility of our findings in this work, we conduct our experiments over both ToolBench and ToolAlpaca, which differ in API quality and instruction requirements.

 


\subsection{Data Quality}
The ever-increasing dependence on data for training large models has paved a line of work that analyzes the effect of data quality on fine-tuning models. Findings show that 
a small but high-quality dataset can be highly effective for fine-tuning
a relatively small model, surpassing the performance of a larger model.
For example, Phi \cite{gunasekar2023textbooks, li2023textbooks} explored code generation tasks and prompted GPT-4 to assess the educational value of coding examples. They demonstrated that a small number of high-quality and diverse examples are sufficient to reach good quality of code generation.
In the realm of instruction tuning, \citet{li2024selfalignment} suggest employing self-augmentation and self-curation to iteratively improve the set of instructions used for instruct-tuning an LLM. 
LIMA \cite{zhou2024lima} considers the broader picture of data quality, and show that as few as 1000 high-quality examples can be sufficient for training an instruction-following model.

Our work differs from these studies in that it applies to the regime of tool usage. It can be seen as additional evidence reinforcing the prevailing ``less is more'' trend, proving the importance of data quality in this regime.

\section{Task Setup}
\label{sec_task_components}
Tool-using LLMs are expected to behave as follows.
Given a set of tools $T = \{t_1, ..., t_n\}$, represented as API functions, and an instruction query $q$, a model is required to plan a call sequence $S = (t'_1, ..., t'_k)$, based on $T$, that would obtain information, or perform actions, needed to address $q$. Based on the responses obtained after performing the call sequence (using an external API invoker), the model then generates a final response $r$ that responds to $q$. The primary method for model evaluation is based on calculating the \textit{pass rate}, which measures the proportion of instances that successfully addressed their instructions, i.e., a predicted $r$ responded to $q$ adequately (explained further in \S\ref{sec_extrinsic_eval}).

As mentioned in Section \ref{sec_related_work}, the prominent datasets created for training and testing tool-using models were created synthetically with the assistance of LLMs. Specifically, we utilize the ToolBench \cite{qin2024toolllm} and ToolAlpaca \cite{tang2023toolalpaca} datasets. \autoref{tab:dataset_summary} summarizes their characteristics. The main practical differences are the quality of the APIs (i.e., the documentation clarity and uniformity of ToolBench is inferior to that of ToolAlpaca), and the number of tools required to respond to a query instruction (ToolBench might require several calls to unrelated tools, while ToolAlpaca requires calling a maximum of two related tools). As presented later in this work, these two differences strongly reflect on the overall quality of the respective datasets.

\begin{table}[ht]
\centering
\resizebox{\columnwidth}{!}{%
\begin{tabular}{lcc}
\toprule
\textbf{Characteristic} & \textbf{ToolBench} & \textbf{ToolAlpaca} \\
\midrule
API source & real-world & synthesized w/GPT \\
\# available APIs & 16K & 2.3K \\
\# of training instances & 125K & 4.2K \\
\# required API calls per instance & 1-5 & 1-2 \\

\bottomrule
 \end{tabular}%
}
\caption{Summary of relevant dataset characteristics.}
\label{tab:dataset_summary}
\end{table}

\paragraph{Problem statement.}


Our primary focus is on evaluating and improving data quality, and to show its effect on model performance in tool-using LLMs. Following a similar line of research, we hypothesize that a small quantity of high-quality training data is preferred over a large quantity of lower-quality data. To demonstrate this, we first define intrinsic quality criteria for the data (\S\ref{sec_intrinsic_eval_framework}) and implement automated metrics accordingly (\S\ref{sec_intrinsic_eval_metrics}). We additionally propose an alternative data quality appraisal method using in-context evaluation (\S\ref{sec_ice}). Finally, we filter out the lower-quality data from datasets using our automated metrics, and analyze the effect of the improved data quality on model performance (\S\ref{sec_extrinsic_eval}). \looseness=-1


\section{Intrinsic Quality Evaluation}
\label{sec_intrinsic}
\subsection{Quality Criteria}
\label{sec_intrinsic_eval_framework}
We set out to understand what makes an instance of data high quality, specifically for training tool-using LLMs. The criteria we discuss pertain to both the query instruction and the API call sequence of a data instance.\footnote{We considered other properties that were eventually excluded from our framework, such as diversity and syntax validity. See Appendix \ref{app:additional_dimenisions} for more details.}


\begin{table*}[ht]
\centering
\begin{tabular}{|c|c|}
\hline
\textbf{Synthetic Instruction} & \textbf{Error Type} \\
\hline
\parbox{0.8\textwidth}{\vspace{1ex}\small{I'm curious about \textbf{a famous actor's} career. Can you provide details about their filmography, including their best-known titles and streaming availability on Netflix, Hulu, and Prime Video? Also, share some interesting facts about the actor.}\vspace{1ex}} & \small{Low Specificity} \\
\hline
\parbox{0.8\textwidth}{\vspace{1ex}\small {As a language enthusiast, I'm always eager to learn new languages. Can you help me explore the possible translations between Russian, Japanese, and Arabic? \textbf{Additionally, I would like to obtain a list of available language codes for future reference}.}\vspace{1ex}} & \small{Low Coherence} \\
\hline
\parbox{0.8\textwidth}{\vspace{1ex}\small {I need to \textbf{create} a temporary email address with the domain 'example.com'. Once created, I want to fetch the latest message from this email address. \\ Given APIs: [Get list of domains for email, Get message by message ID]}\vspace{1ex}} & \small{Unsolvable} \\
\hline
\end{tabular}
\caption{Examples of synthesized instructions, \textbf{highlighted} with errors involving our defined properties.}
\label{tab:examples_instruction}
\end{table*}

\begin{table*}[ht]
\centering
\resizebox{\textwidth}{!}{%
\begin{tabular}{|c|c|c|}
\hline
\textbf{Synthetic Instruction} & \textbf{API-Call within Sequence} & \textbf{Error Type} \\
\hline
 \parbox{0.5\textwidth}{\vspace{1ex}\small {Can you create a shield logo for my friend's blog? The name of the blog is `The Creative Mind'.}\vspace{1ex}} & \parbox{0.25\textwidth}{\small\texttt{generate\_shield(name=None)}} & \small{Missing Parameter} \\
 
\hline
\parbox{0.5\textwidth}{\vspace{1ex}\small{I need to fetch the current weather conditions for a specific location. Can you help me by providing the address and geocoordinates of the location?}\vspace{1ex}} & \parbox{0.25\textwidth}{\small \texttt{geocode(address="San Francisco")}  \\ ...}

 & \small{Hallucinated Parameter} \\
\hline
\end{tabular}%
}
\caption{Examples of synthesized API-call sequences for respective instructions, with incorrect parameters.}
\label{tab:examples_api}
\end{table*}

\subsubsection{Instruction Properties}


In our setting, an instruction is a free-form text of one-to-a-few sentences that describes a user requirement.
An instruction can contain more than one request, likely implying the need for several tool invocations. The following properties in the instruction demand validation (examples in \autoref{tab:examples_instruction}):

\paragraph{Specificity.} All the required details are present in the instruction for the LLM to be able to fulfill the user requests.\vspace{-3pt} 
\paragraph{Coherence.} The requests within the instruction are logically related, and the order of requests makes sense for a real-world use case.\vspace{-3pt}
\paragraph{Solvability.} The requests within the instruction can be addressed by the given API tools.

\subsubsection{API-Call Sequence Properties}
Apart from the instruction, given as input to a model, the other vital component of a training instance is the ground-truth output used for training (or evaluating) a model. In our setting, this is the sequence of API calls that the model is expected to infer. 
We define the following properties for API-call sequence correctness (see \autoref{tab:examples_api} for examples): \looseness=-1
\paragraph{Parameter alignment.} The parameter values in each of the API calls are correctly extracted or inferred from the instruction, there are no missing or hallucinated parameter values.

\paragraph{Sufficiency.} The API-call sequence applies to all required actions for the instruction's requests.

\paragraph{Minimality.} The API-call sequence would address all the instruction requirements with a minimal number of API calls. No unnecessary or redundant API calls are included in the sequence.

\subsection{Manual Annotations}
\label{sec_manual_annotation}

The six intrinsic properties defined above specify the desired qualities for data instances of tool-using LLMs. Existing datasets do not always abide by these quality criteria, especially when they are collected synthetically and do not go through a cleaning phase. We inspect such noisy data by preparing annotation guidelines with respect to the criteria, and annotating accordingly. Specifically, we methodically\footnote{Annotation process and agreement in Appendix \ref{appendix_manual_annotation}.} annotated 50 (instruction, API sequence) pairs from each of the training sets of ToolBench \cite{qin2024toolllm} and ToolAlpaca \cite{tang2023toolalpaca}, as well as a large portion of the ToolBench test set ($\sim$700 instances).\footnote{We did not review ToolAlpaca's test set since it is already manually verified.} Each of the criteria is marked either as valid or invalid for each of the annotated instances. 
The annotated data is used in later sections for analyses and experiments.
\begin{table*}[ht]
\centering
\renewcommand{\arraystretch}{1.2} 
\setlength{\tabcolsep}{5pt} 
\resizebox{0.8\textwidth}{!}{%
\begin{tabular}{clcccc|ccccc}
\toprule
& & \multicolumn{4}{c}{\textbf{ToolBench Dataset}} & \multicolumn{4}{c}{\textbf{ToolAlpaca Dataset}} \\
\cline{3-6} \cline{7-10}
& \textbf{Quality Criterion} & \textbf{Accuracy} & \textbf{Prec.} & \textbf{Rec.} & \textbf{F1} & \textbf{Accuracy} & \textbf{Prec.} & \textbf{Rec.} & \textbf{F1} \\
\hline
\multirow{4}{*}{\rotatebox[origin=c]{90}{Instruction}} & Specificity & 0.74 & 0.70 & 0.84 & 0.76 & 0.88 & 0.75 & 0.86 & 0.80 \\
& Coherence & 0.82 & 0.62 & 0.77 & 0.69 & 0.98 & 0.50 & 1.00 & 0.66 \\
& Solvability & 0.90 & 0.70 & 0.78 & 0.74 & 0.92 & 0.75 & 0.50 & 0.60 \\ \cline{2-10}
& Instruction Correctness & 0.72 & 0.72 & 0.90 & 0.80 & 0.86 & 0.80 & 0.84 & 0.82 \\
\hline \hline
\multirow{4}{*}{\rotatebox[origin=c]{90}{API Call Seq.}} & Parameter Alignment & 0.70 & 0.63 & 0.92 & 0.74 & 0.76 & 0.74 & 0.80 & 0.77 \\
& Sufficiency & 0.78 & 0.64 & 0.60 & 0.62 & 0.88 & 0.80 & 0.50 & 0.62 \\
& Minimality & 0.76 & 0.95 & 0.63 & 0.76 & 0.86 & 0.88 & 0.57 & 0.70 \\ \cline{2-10}
& Sequence Correctness &  0.82 &  0.83 &  0.94 &  0.88 &  0.76 &  0.70 &  0.85 &  0.80 \\
\hline \hline
& \multicolumn{1}{l}{Overall Correctness} &  0.86 &  0.89 &  0.95 &   0.92 &  0.76 &  0.74 &  0.90 &  0.81 \\
\bottomrule
\end{tabular}%
}

\caption{Validation results of the automated metrics for each criterion, in the ToolBench and ToolAlpaca datasets. Coarse-grained correctness considers combined correctness over specific criteria. Note that precision, recall and F1 are measured w.r.t.\ a label that is positive when an error occurs, so e.g., recall means the amount of errors caught. 
}

\label{tab:intrinsic_metrics}
\end{table*}

\subsection{Automated Metrics}
\label{sec_intrinsic_eval_metrics}

Although manual assessment of data is preferred for its reliability, it is labor-intensive and therefore not scalable or practical. We propose automatic metrics for the intrinsic quality criteria defined above.
The metrics are based on \evaluatorModel,\footnote{Throughout the paper, we use \texttt{gpt-3.5-turbo-0613}.} which is tasked to determine the validity of each criterion as a binary decision.

For the dimensions of Specificity, Coherence and Parameter alignment, direct annotation with \evaluatorModel{} proved to be challenging. That is, simply asking the model to validate the property in a natural language instruction did not yield sufficient decisions (see Appendix \ref{app:unused_prompts}). 
Thus, we transformed the direct annotation tasks into traditional NLP tasks, on which \evaluatorModel{} performed better.\looseness=-1

\vspace{-5pt}

\paragraph{Specificity.}
Validating the specificity of requests is modeled as
an \textit{extraction} task. \evaluatorModel{} is tasked to infer the details required for a given request, and then extract the available values from the instruction, or mark a parameter as \texttt{\#missing}. We then compute a proxy score for specificity: 1 if all parameters were successfully extracted from the instruction, and 0 otherwise.

\vspace{-5pt}


\paragraph{Coherence.}
We adopt the concept of \textit{next sentence prediction} to assess coherence. The instruction is split into sentences, and \evaluatorModel{} determines if each subsequent sentence logically follows the previous one. We set a coherence score as 1 if all sentence pairs are judged logically connected, and 0 otherwise.\looseness=-1

\vspace{-5pt}

\paragraph{Parameter alignment.}
\evaluatorModel{} first extracts parameters (as in specificity), and then compares it to the ground truth parameter values.

\paragraph{Solvability, Sufficiency \& Minimality.} The remaining criteria use direct instructions to \evaluatorModel. The prompts used are provided in Appendix \ref{app:prompts}.

\begin{table*}[h]
\centering
\renewcommand{\arraystretch}{1.4} 
\resizebox{\linewidth}{!}{
\begin{tabular}{|l| c|c|c||c|  c|c|c||c  ||c|}
\hline
\multirow{3}{*}{\textbf{Dataset}} & \multicolumn{4}{c|}{\textbf{Instruction}} & \multicolumn{4}{c||}{\textbf{API-Call Sequence}} & \multirow{2}{60pt}{\textbf{Inst.\ \& Seq.\ Overall}} \\

& \textbf{Specificity} & \textbf{Coherence} & \textbf{Solvable} & \textbf{Overall} & \textbf{Param. Alignment} & \textbf{Sufficiency} & \textbf{Minimality} &  \textbf{Overall} & \\

\hline

\multirow{1}{*}{\textbf{ToolBench}}  & 20.4\% & 22.1\% & 18.2\%  & 47.3\% & 47.9\% & 33.6\% & 45.1\% & 74.4\% & \multirow{1}{*}{84.0\%} \\


\hline
\multirow{1}{*}{\textbf{ToolAlpaca}}  & 17.5\% & 4.1\% & 12.7\%  & 27.2\% & 33.1\% & 13.6\%  & 15.9\% & 35.5\% & \multirow{1}{*}{44.8\%} \\
\hline
\end{tabular}}
\caption{Percentage of instances containing errors in each dimension, according to our automated methods, in the train sets of the examined datasets. This analysis is done on 125K examples in ToolBench and 4.2K in ToolAlpaca.}
\label{tab:data_statistics}
\end{table*}

\subsubsection{Evaluation of Automated Metrics}

Using the manually annotated data (described in \S\ref{sec_manual_annotation}), we conduct an assessment of the automatic metrics proposed. For each of the ToolBench and ToolAlpaca datasets, the 50 annotated instances are compared against the automatically produced values, producing measures of accuracy (agreement), precision, recall and F1 score. We treat instances marked as incorrect instances as positive labels, since we aim to identify and filter erroneous instances. 

We conduct a coarser-grained evaluation of the criteria, assessing \textbf{Instruction Correctness} as incorrect if any instruction criterion is wrong, and \textbf{Sequence Correctness} as incorrect if any API-call sequence criterion is wrong. \textbf{Overall Correctness} aggregates all six criteria similarly.


Results are presented in \autoref{tab:intrinsic_metrics}. Given that our main objective is to identify and filter out incorrect data samples, our emphasis is on achieving high recall. This objective is largely met across most criteria in both datasets. In the Overall Correctness assessment, which aggregates all criteria, we observe high recall and precision, demonstrating a strong alignment of the automated metrics with human judgment. This approach thus offers a reliable mechanism to identify problematic data instances.\looseness=-1

\subsubsection{
Quality of Datasets
}
\label{sec:intrinsic_analysis}
Table \ref{tab:data_statistics} presents the percentage of instances containing errors in the train sets of both ToolBench and ToolAlpaca, as determined by the automated metrics. These statistics provide insights into the quality of the data in each dataset. In the ToolBench dataset we observe a much higher percentage of errors across most quality criteria, when compared to ToolAlpaca. This difference may be attributed to (1) the complexity of instructions in ToolBench, which can 
require several (up to 5) API calls; (2) real-world APIs used in ToolBench, where the API documentation is not always clear, resulting in incorrectly generated instructions and API-call sequences. 
Notice that in both datasets, over $33\%$ of instances have parameter alignment errors. Such an error means that one of the core requirements of a tool-using model -- identifying parameters correctly -- is misleadingly learned in more than a third of the cases, due to wrong training examples.
%
Some anecdotal examples of incorrect instructions found by our metrics can be seen in Appendix \ref{app:qualitative_examples}.

We further explore the relationship between quality criteria within the datasets in Appendix \ref{app:dimensions_relations}.



\section{In-Context Evaluation (ICE) as an Alternative Data Measurement}

\label{sec_ice}
Using intrinsic evaluation, we have defined an intuitive and straightforward approach to identify low-quality data instances based on human understanding of data correctness. However, assessing the ``educational'' value of an instance, i.e., its contribution to the learning process of a model, is a complex task.
In addition, the intrinsic evaluation metrics proposed rely on prompting a powerful LLM, which can become costly on large datasets. To address these challenges, we propose In-Context Evaluation (ICE) as an alternative automatic approach for assessing data quality.

Recent studies found a connection between in-context learning and fine-tuning, demonstrating that language models implicitly perform gradient descent when dealing with in-context tasks \cite{von2023transformers,dai-etal-2023-gpt}. Motivated by this insight, we seek to evaluate the educational value of each data instance by measuring the performance of in-context learning using the specific instance as a one-shot example. 


\subsection{Setup}

To construct the in-context task for external tool use, we prepare a set of 10 human-written APIs, denoted by $A$, with simple accompanying documentation.
In addition, we hand-craft a set of 7 test query instructions, \textsc{Test}, where each such example contains a natural language instruction and an expected API-call sequence, from the APIs in $A$, that would address the instruction.
For each evaluation instance, we insert an in-context example, $x$, which consists of an instruction and API-call sequence from the training dataset (i.e., ToolBench or ToolAlpaca).
$x$ follows the structure of the test examples.
We then formulate a prompt for the LLM that we aim to train, that asks to generate responses for the 7 test cases. In particular, the prompt includes (1) task instructions,
(2) the API documentation of $A$, (3) the training instance, $x$, given as a one-shot example, (4) the 7 testing instructions of \textsc{Test}.

The prompt is given to an LLM we aim to train: LLaMA-7B for ToolBench or Vicuna-7B for ToolAlpaca. We analyze its response, that should include the 7 API-call sequences of \textsc{Test}.
The responses for the test instructions are evaluated against the ground truth (using Levenshtein similarity \cite{levenshtein1966binary}, expecting an exact match for API-call sequences). The final ICE score for $x$ is the average over the 7 test examples, interpreted as a measure of the educational value of~$x$. We provide the full prompt and the precise way we compute
the ICE score in Appendix \ref{app:ice}.


\subsection{Analysis}

\begin{figure}[b!]
  \centering
  \begin{subfigure}{0.5\textwidth}
    \centering
    \includegraphics[width=\textwidth,height=0.45\textwidth]{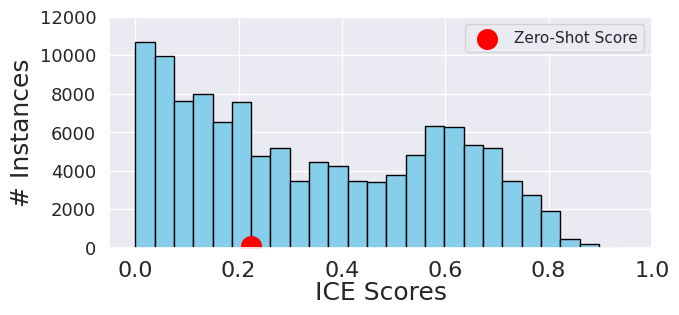}
    \caption{ToolBench}
    \label{fig:ice_distribution_bench}
  \end{subfigure}
    \begin{subfigure}{0.5\textwidth}
    \centering
    \includegraphics[width=\textwidth,height=0.45\textwidth]{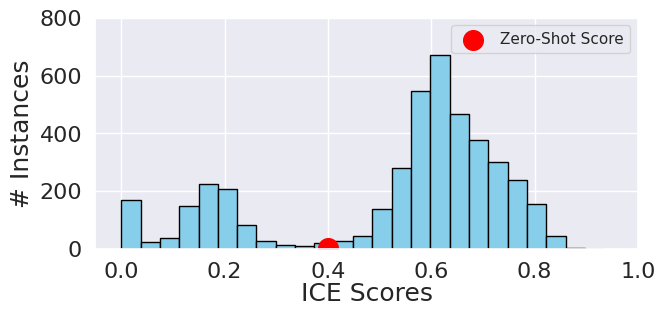}
    \caption{ToolAlpaca}
    \label{fig:ice_distribution_alpaca}
  \end{subfigure}
  \caption{Distribution of ICE scores. Most instances in ToolAlpaca are beneficial as the one-shot in-context example. ToolBench instances are not as effective.}
  \label{fig:ice_distribution}
\end{figure}


\begin{figure}[t!]
    \centering
    \includegraphics[width=\linewidth,height=4cm]{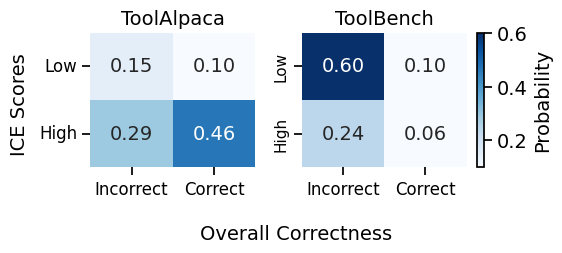}
    \caption{Confusion matrices comparing ICE scores and human Overall Correctness scores.}
    \label{fig:ice_corr}
\end{figure}

\paragraph{Score distribution.} We present ICE scores for both datasets in Figure \ref{fig:ice_distribution}. Interestingly, the ICE scores distribution in ToolAlpaca exhibits 
bimodal distribution, which suggests the presence of two types of examples: one with higher ICE scores, which we expect to correlate with good-quality examples, and another with lower ICE scores, which is expected to lean towards low-quality examples. 
The majority of instances in ToolAlpaca have relatively high ICE scores -- indicating high overall dataset quality. In contrast, most samples in ToolBench have low ICE scores, suggesting that the overall data quality in this dataset may be lower compared to ToolAlpaca. This observation is consistent with the analysis presented using the intrinsic evaluation in Section \ref{sec:intrinsic_analysis}.\looseness=-1

\paragraph{Correlation to human-defined criteria.}

ICE is a model-driven assessment method that may not necessarily align with human-defined correctness criteria. To investigate the relationship between ICE approach and human-defined criteria, we divide the datasets into low and high ICE scores using a threshold of 0.5. We then generate confusion matrices between ICE scores and human \textit{Overall Correctness} scores.
As seen in Figure \ref{fig:ice_corr},  
%
ICE scores correlate with human-defined correctness to some extent,
showing it is a sensible metric and can be beneficial as an alternative method for filtering data. On the other hand, this correlation is far from perfect, showing that ICE is inherently different from human-prescribed correctness.
In Section \ref{sec_extrinsic_eval} we test ICE both as an alternative and as a complementary filtering technique to human-defined correctness. 

\section{Extrinsic Evaluation}
\label{sec_extrinsic_eval}

\renewcommand{\arraystretch}{1.2}
\begin{table*}[h]
\centering
\begin{tabular}{clccc|ccc}
\cmidrule[\heavyrulewidth]{2-8}
& \multirow{2}{*}{\textbf{Fine-tune Set}} & \multicolumn{3}{c}{\textbf{ToolBench}} & \multicolumn{3}{c}{\textbf{ToolAlpaca}} \\
\cline{3-8}
& & \textbf{Size} & \textbf{Pass Rate} & \textbf{95\% CI} & \textbf{Size} & \textbf{Pass Rate} & \textbf{95\% CI} \\

\cline{2-8}

\textcolor{gray}{1} & Random Sample & 10K & $0.35$ & $$\small (0.31, 0.39)$$ & 2K & $0.48$ & $$\small (0.38, 0.58)$$ \\

\textcolor{gray}{2} & Low ICE & 10K & $0.24$ & $$\small (0.20, 0.28)$$ & 2K & $0.48$ & $$\small (0.38, 0.58)$$ \\

\textcolor{gray}{3} & High ICE & 10K & $0.43$ & $$\small (0.38, 0.47)$$ & 2K & $0.54$ & $$\small (0.44, 0.64)$$ \\

\textcolor{gray}{4} & High Instruction & 10K & $0.49$ & $$\small (0.44, 0.53)$$ & 2K & $0.52$ & $$\small (0.42, 0.62)$$ \\

\textcolor{gray}{5} & High Instruction + Seq & 10K & $0.52$ & $$\small (0.47, 0.56)$$ & 2K & $0.54$ & $$\small (0.44, 0.64)$$ \\

\textcolor{gray}{6} & High Instruction + Seq + ICE & 10K & $0.54$ & $$\small (0.49, 0.58)$$ & 2K & $0.55$ & $$\small (0.45, 0.65)$$ \\

\cline{2-8}

\textcolor{gray}{7} & Original & 73K$^\dagger$ & $0.45$ & $$\small (0.40, 0.49)$$ & 4.2K & $0.56$ & $$\small (0.46, 0.66)$$ \\ 

\cmidrule[\heavyrulewidth]{2-8}

\end{tabular}
\caption{Extrinsic evaluation results with confidence intervals, and the size of the training sets. 
By filtering out low-quality training instances, the models perform significantly better than (in ToolBench) or as good as (in ToolAlpaca) the original models that use a much larger unvalidated training set.{ \small $^\dagger$ Although there are 125K instances in the released dataset, the model published in the original paper was trained on a subset of 73K instances.}}
\label{tab:extrinsic_results}
\end{table*}

In this section, we validate our main claim that fine-tuning a tool-using LLM with less higher-quality data can lead to better performance of the model on the task, compared to a more noisy dataset. We use both intrinsic metrics and ICE to create training sets of varying quality, and compare the results of training with the different sets.
 
\paragraph{Training setup.}
We follow the general setup used by the ToolAlpaca \citep{tang2023toolalpaca} and ToolBench \citep{qin2024toolllm} benchmarks. Specifically, we fine-tune Vicuna-7B \cite{chiang2023vicuna} for ToolAlpaca, and LLaMA-7B \cite{touvron2023llama} for ToolBench, both using LoRA \cite{hu2022lora} (see Appendix \ref{app:training} for more details).

We use the following train sub-sets from each model's respective benchmark training sets:

\begin{itemize}
    \setlength{\itemsep}{1pt}
    \setlength{\itemindent}{-7pt}
    \setlength{\parskip}{0pt}
    \setlength{\parsep}{0pt}
    \item \textbf{Random Sample}: uniform random subset.
    \item \textbf{High Instruction}:  uniform sample of instances with all three instruction criteria intact.
    \item \textbf{High Instruction + Seq}: uniform sample of instances with all six criteria intact.
    \item \textbf{Low ICE}: instances with the lowest ICE scores.
    \item \textbf{High ICE}: instances with the highest ICE scores.
    \item \textbf{High Instruction + Seq + ICE}: instances with all six criteria intact and high ICE scores. 
    \item \textbf{Original}: the full original training set.

\end{itemize}

Each fine-tuned model is evaluated using \textbf{pass rate}, which is an extrinsic evaluation procedure used in both benchmarks.\footnote{In ToolAlpaca this metric is referred to as ``overall accuracy'', although it conveys the same concept.} This measures the proportion of instances in which the resulting API-call sequences and responses adequately address their respective instruction query. See Appendix \ref{app:evaluation} for more details on the evaluation procedure.

\paragraph{Test sets.}
For ToolAlpaca we use the original test set, as it is created with human annotation. It consists of 100 instructions of simulated tools that were not part of the training tool set. ToolBench test set was created using LLMs and was not manually validated. We inspected 674 examples, as detailed in Appendix \ref{appendix_manual_annotation}. For instances of low quality, we either rectified them (e.g., manually adding a missing parameter value), or discarded them. The resulting test set contains 420 high-quality examples.\footnote{This test set is available in the supplementary material.} 







\subsection{Main Results}
\label{sec:main_results}
Results are presented in \autoref{tab:extrinsic_results}, where the training sub-sets are fixed to size 10K for ToolBench and 2K for ToolAlpaca. The results demonstrate the impact of training data quality on model performance.

When comparing to a model fine-tuned on a random subset of the original training data (row \textcolor{gray}{1}), all methods of filtering low-quality instances (rows \textcolor{gray}{3}-\textcolor{gray}{6}) are clearly beneficial.
Moreover, when fine-tuning models with \textit{much smaller} high-quality sub-sets (rows \textcolor{gray}{3}-\textcolor{gray}{6}), performance is comparable or superior to models fine-tuned on the \textit{full} original training sets (row \textcolor{gray}{7}). Consistent with the findings on the ToolBench dataset's lower overall quality (\S\ref{sec_intrinsic} and \S\ref{sec_ice}), results indicate improved model performance with a high-quality subset, comprising only $\sim$14\% of the original dataset's size (row \textcolor{gray}{6}).

Comparing the intrinsic metrics to the ICE method, we find that the former is a better mechanism for filtering training data (row \textcolor{gray}{3} vs. \textcolor{gray}{5}). Using both techniques together can be marginally better (row \textcolor{gray}{5} vs. \textcolor{gray}{6}). Another insight to consider is that taking data with low ICE scores (row \textcolor{gray}{2}) is indeed harmful to model performance, further reinforcing that the method is valuable despite its partial agreement with intrinsic human-defined criteria (\S\ref{sec_ice}).

In ToolAlpaca, the gaps are less pronounced than in ToolBench, likely influenced by: (1) the higher quality of the original dataset, (2) smaller original training set, causing the filtered datasets to be too small, (3) smaller test set, only 100 instances. Nonetheless, the trend still exists (albeit being within the confidence intervals). This, combined with the intrinsic assessment of Table~\ref{tab:data_statistics}, provides encouraging evidence for the effectiveness of our methods, even for this smaller-scale dataset.

\begin{figure}[t]
  \centering
  \includegraphics[width=\linewidth,height=0.72\linewidth]{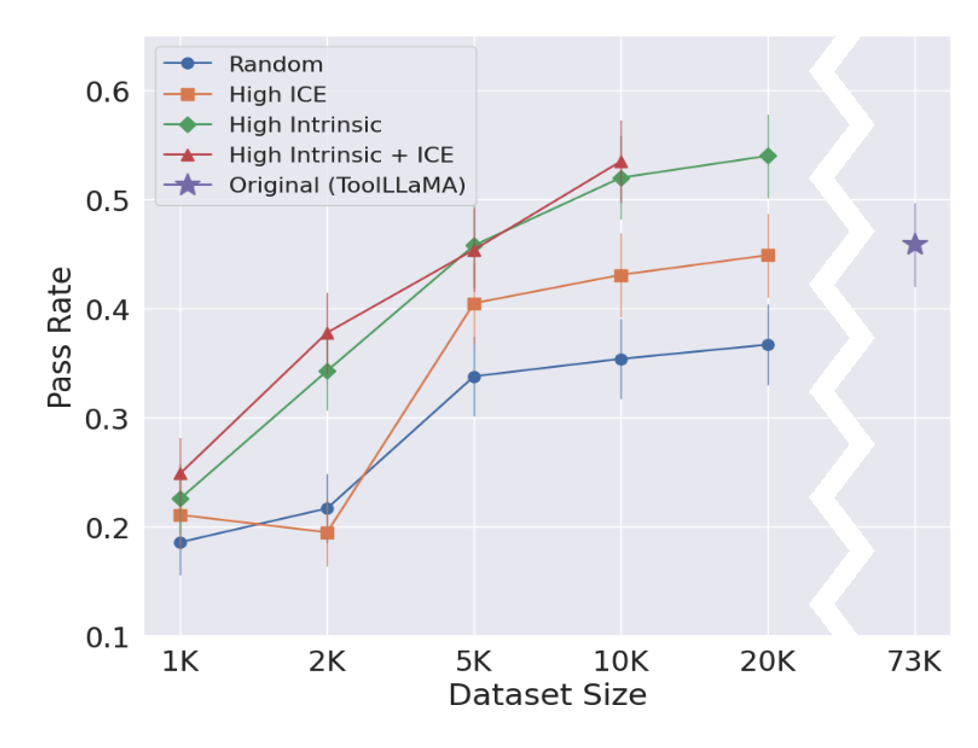}
  \caption{Pass rate results in ToolBench when using train sets with different sizes and filtration methods.}
  \label{fig:varying_sizes}
\end{figure}

\subsection{Data Scaling Analysis}
To further explore the effects of training tool-using LLMs with high-quality data, we analyze the performance of models when fine-tuning with \textit{different sizes} of train sets. We focus here on ToolBench, where the impact is more significant and the original training set is larger, and use subsets with sizes ranging from 1K to 20K for the different filtration methods. Results can be found in \autoref{fig:varying_sizes}. As size increases, we observe consistently better performance, with an expected plateau in the largest dataset sizes. Notice that at some point the training datasets have no more high-quality data instances that pass our filters, putting a natural limit on our experiments.  \looseness=-1


\section{Conclusion}
\label{sec_conlusion}
We demonstrated the importance of evaluating the quality of training data for fine-tuning tool-using LLMs. 
We introduce two data-evaluation approaches. The first is a rigorously devised intrinsic quality assessment, for which we implement automated metrics. The second uses in-context evaluation, that measures the educational value of training examples.
While the former method is more explainable and dependable, the latter is computationally cheaper. 
We apply both approaches to filter data instances from two large datasets of differing qualities. The resulting subsets of training data demonstrate comparable or superior quality in terms of model performance, despite their smaller size compared to the original datasets.
Overall, we observe that it is worthwhile to more carefully choose the training data for tool-using LLMs. If investing in better methods of data generation is costly, automatic post-hoc filtration can be a great alternative. \looseness=-1

\section{Limitations}
In this work, we address the quality of data instances, and refrain from overall dataset-level quality criteria, primarily diversity of data. Our focus is on instance-level quality, and we show the advantage of training LLMs with data that is identified as high-quality with instance-level criteria only. Future work can explore the benefits of dataset-level quality criteria as well.

Our experiments span over two popular benchmarks for tool-using LLMs. They are differing in characteristics and quality, and can therefore provide insights that are not benchmark-specific. Nevertheless, conducting our analyses on additional related datasets and LLMs would provide an even more generalized representation of our results.

\bibliography{custom}

\clearpage
\appendix

\section{Intrinsic Evaluation}
\label{app:intrinci}
\subsection{Other Quality Criteria}
\label{app:additional_dimenisions}
We outline here three quality criteria that are commonly addressed in the domain of data quality evaluation, and that we did not include in this work. (1) \textbf{Fluency} is the lexical quality of the text in terms of grammar, spelling, and
style \cite{celikyilmaz2020evaluation}. The reason for omitting this dimension is that the lexical quality of texts generated by powerful LLMs is very high. We found that virtually all instances in an assessment set had highly fluent texts. 
(2) \textbf{Syntax Validity} is whether the function calls and parameter names (not values) in the API-call sequence are valid. Using both manual validation and automatic rule-based lexical matching we found that data generated with \evaluatorModel{} did not exhibit such errors.
(3) \textbf{Diversity} captures how different the data instances are amongst themselves in terms of assortment of requests, tool usage, difficulty, length and other properties. Similarly to other tasks and domains, it is expected that an LLM would learn to generalize better given diverse examples \cite{diversity_in_ml, tale_of_diversity}. We focus on instance-level criteria, and leave dataset-level criteria, such as diversity, for future work.

\subsection{Prompts for Assessment}
\label{app:prompts}
\begin{figure}[t]
  \centering
  \includegraphics[width=\linewidth]{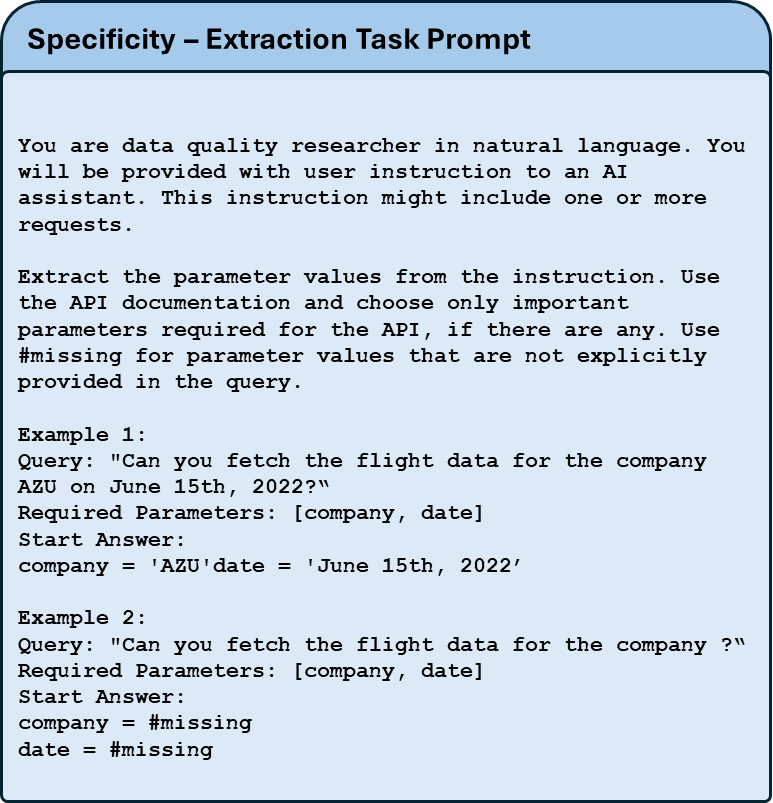}
  \caption{Prompt for instruction \textbf{specificity}, as an extraction task.
  }
  \label{fig:prompt_specificity_task}
\end{figure}

\begin{figure}[h]
  \centering
  \includegraphics[width=\linewidth]{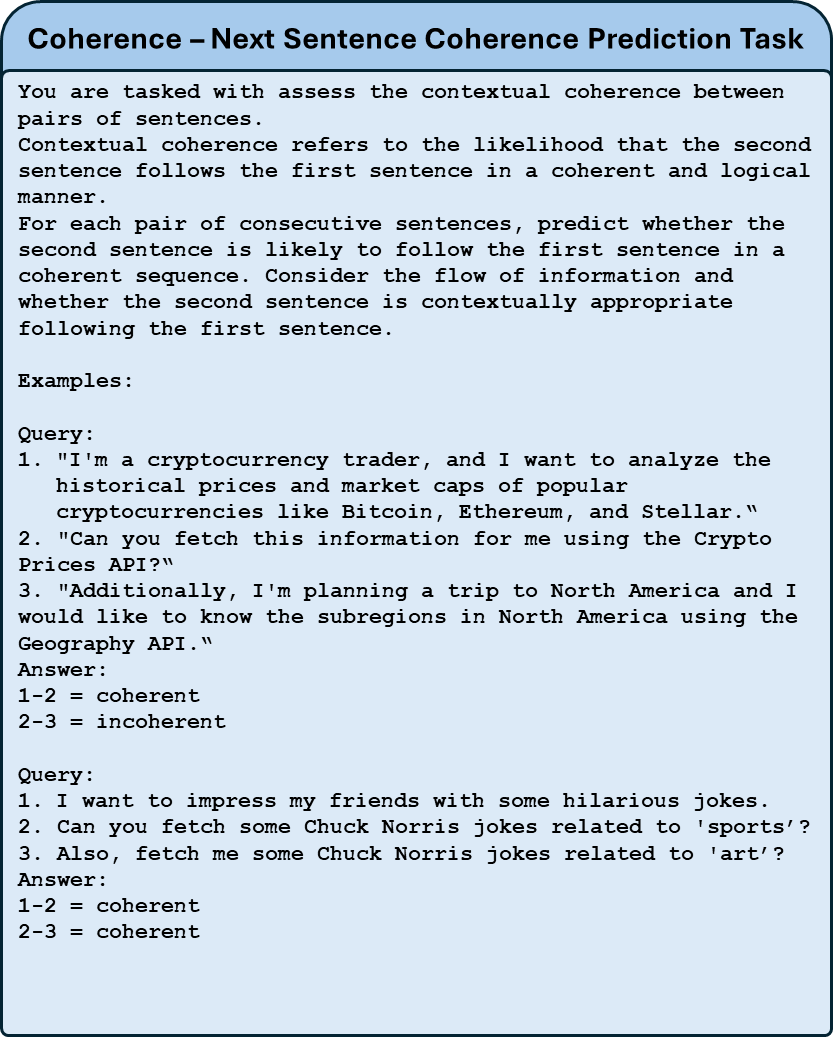}
  \caption{Prompt for instruction \textbf{coherence}, as a next sentence coherence prediction task. 
  }
  \label{fig:prompt_coherence_task}
\end{figure}

\begin{figure}[h]
  \centering
  \includegraphics[width=\linewidth]{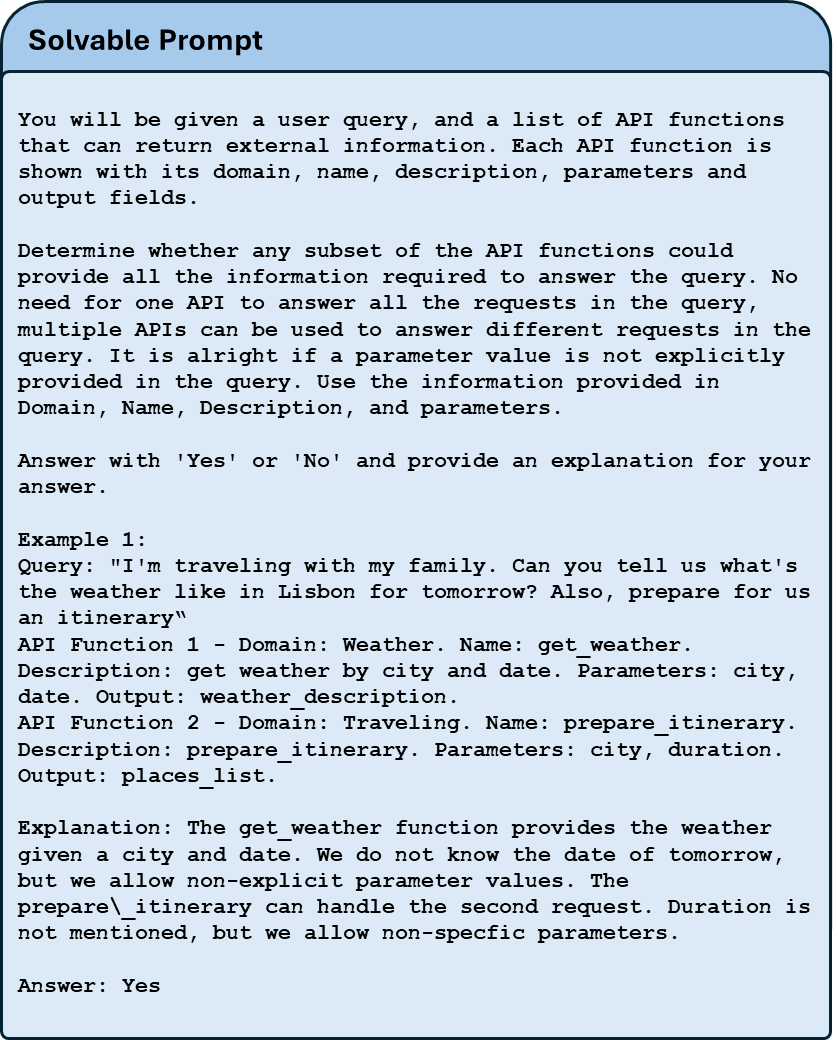}
  \caption{Prompt for instruction \textbf{solvability}. }
  \label{fig:prompt_solvable_direct}
\end{figure}

\begin{figure}[h]
  \centering
  \begin{subfigure}{\linewidth}
    \centering
    \includegraphics[width=0.9\linewidth]{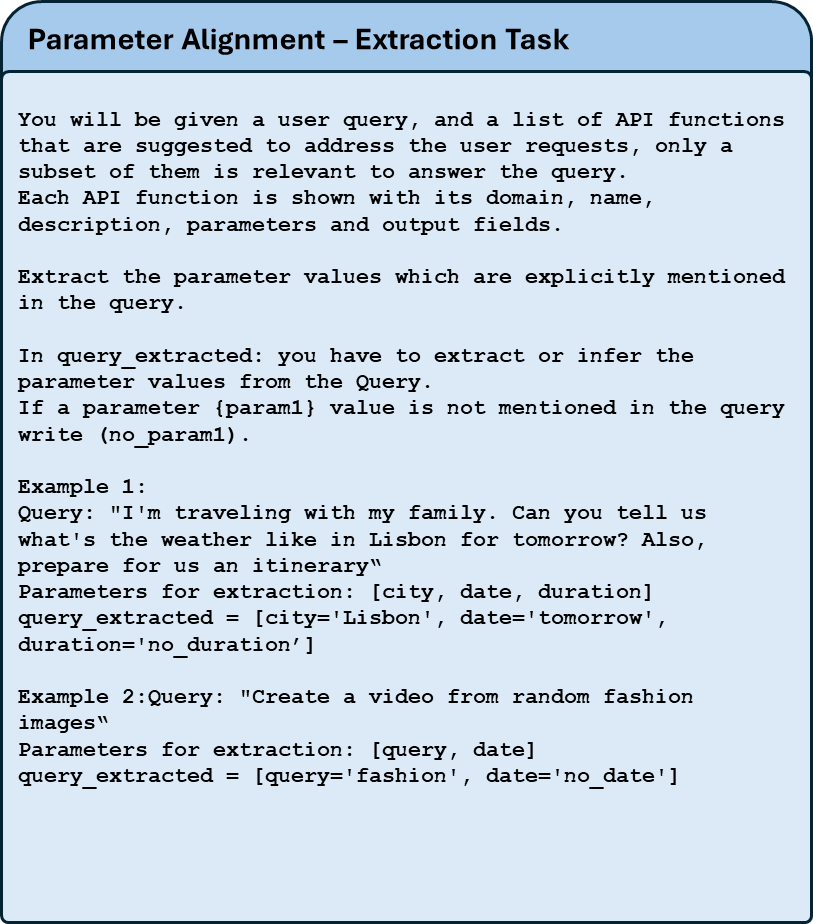}
    \caption{Step 1: parameter value extraction}
    \label{fig:prompt_param_extract1}
  \end{subfigure}
  
  \begin{subfigure}{\linewidth}
    \centering
    \includegraphics[width=0.9\linewidth]{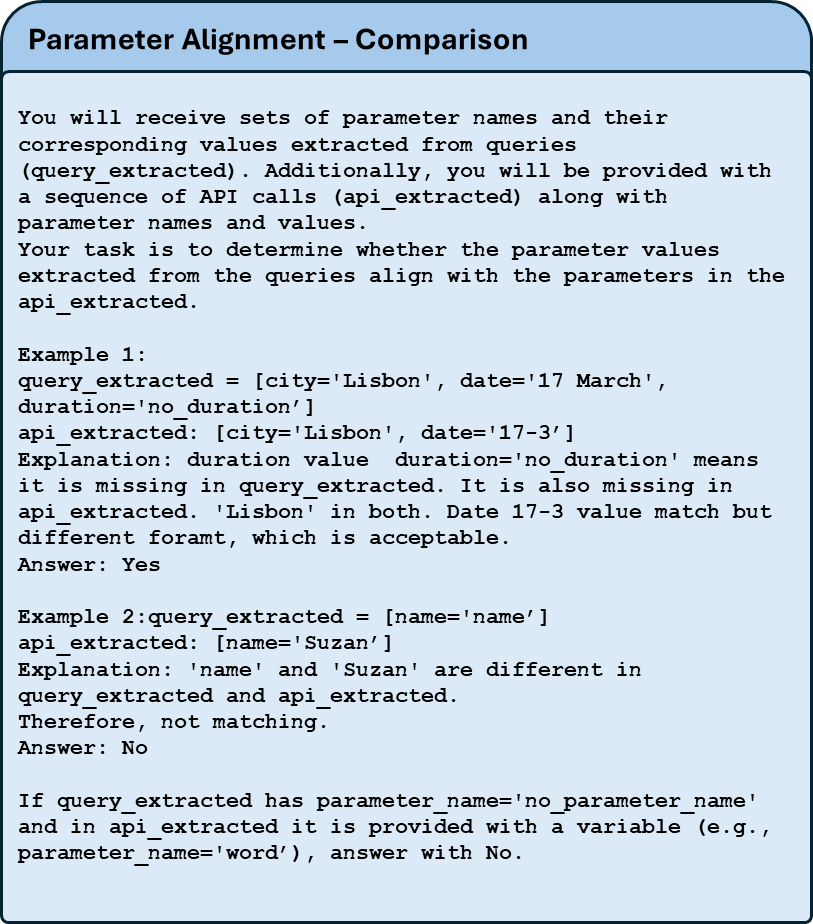}
    \caption{Step 2: comparison}
    \label{fig:prompt_param_compare2}
  \end{subfigure}
  
  \caption{Prompts for assessing \textbf{parameter alignment} in the API-call sequence, as a two-step procedure.}
  \label{fig:prompt_param_task}
\end{figure}

\begin{figure}[h]
  \centering
  \includegraphics[width=\linewidth]{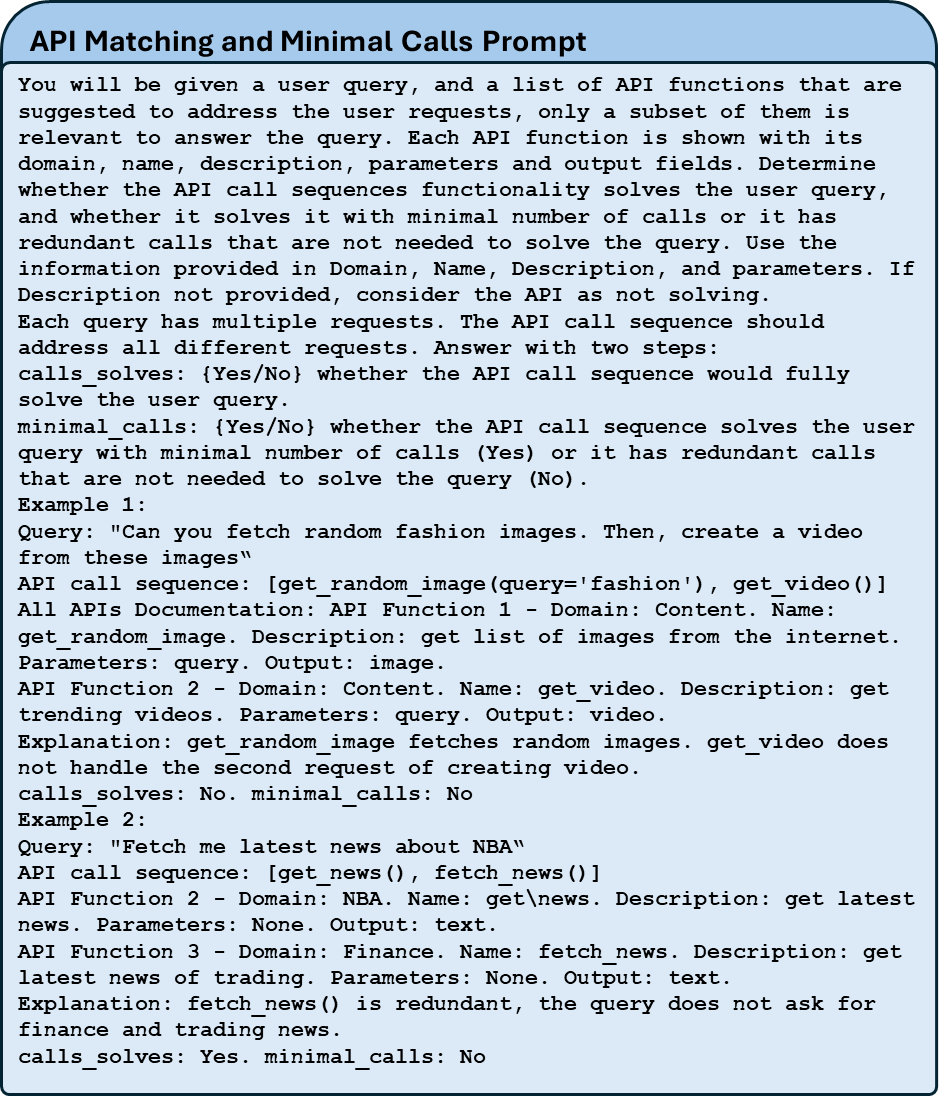}
  \caption{Prompt for \textbf{sufficiency} and \textbf{minimality} of the API-call sequence.
  }
  \label{fig:prompt_api}
\end{figure}

In Figures \ref{fig:prompt_specificity_task}--\ref{fig:prompt_api} we provide the prompts we use for automatically assessing the six human-defined quality criteria, using \evaluatorModel{}.

\subsection{Unsuccessful Prompts}
\label{app:unused_prompts}
\begin{figure}[h]
  \centering
  \includegraphics[width=\linewidth]{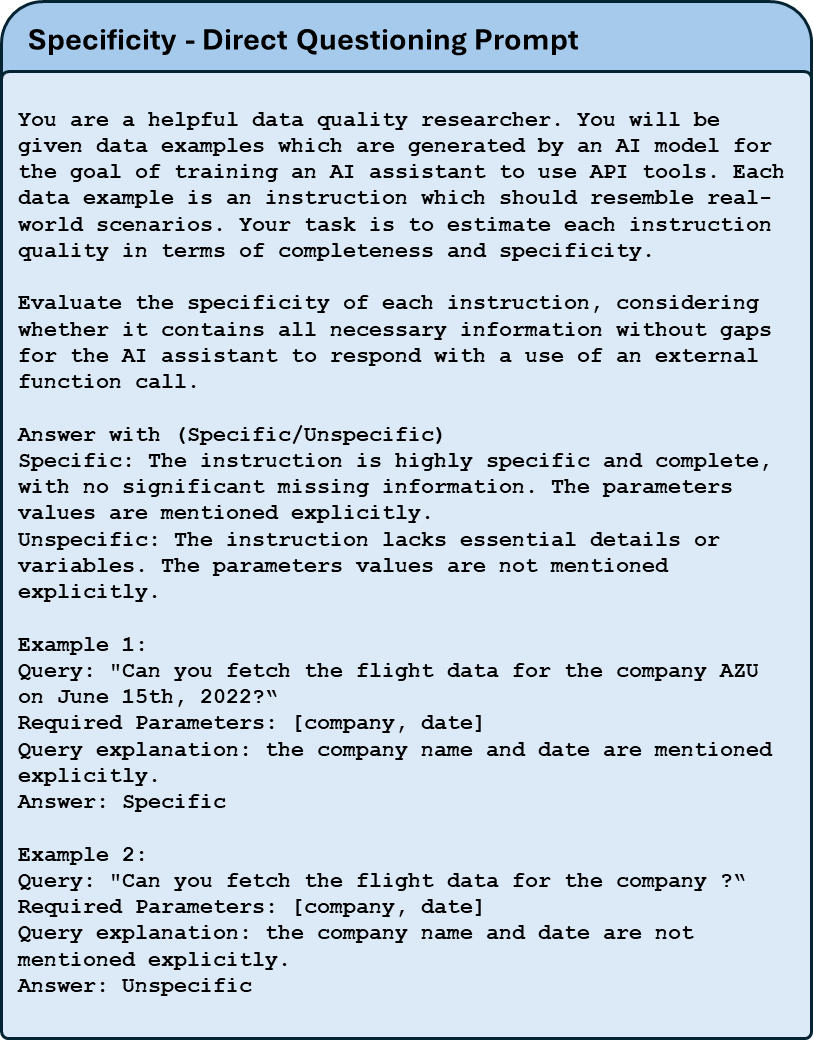}
  \caption{Prompt for \textbf{specificity}, as a direct questioning task.}
  \label{fig:prompt_specificity_direct}
\end{figure}

\begin{figure}[h]
  \centering
  \includegraphics[width=\linewidth]{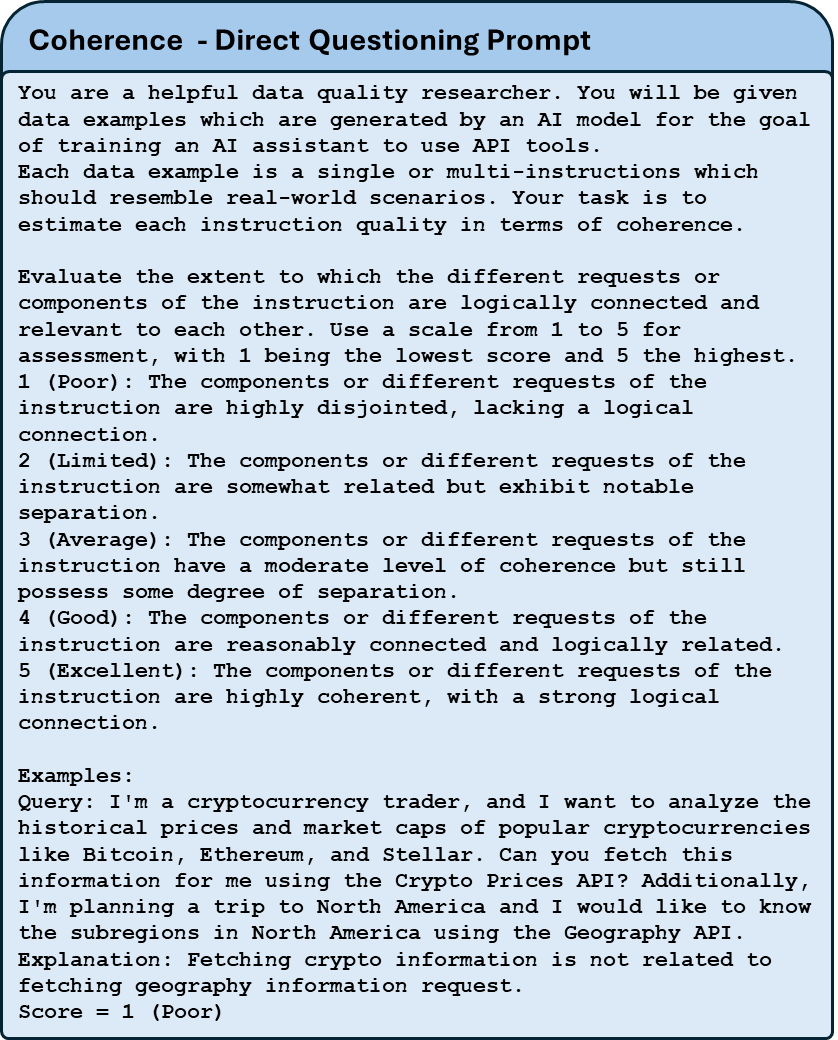}
  \caption{Prompt for \textbf{coherence}, as a direct questioning task.
  }
  \label{fig:prompt_coherence_direct}
\end{figure}

\begin{figure}[h]
  \centering
  \includegraphics[width=\linewidth]{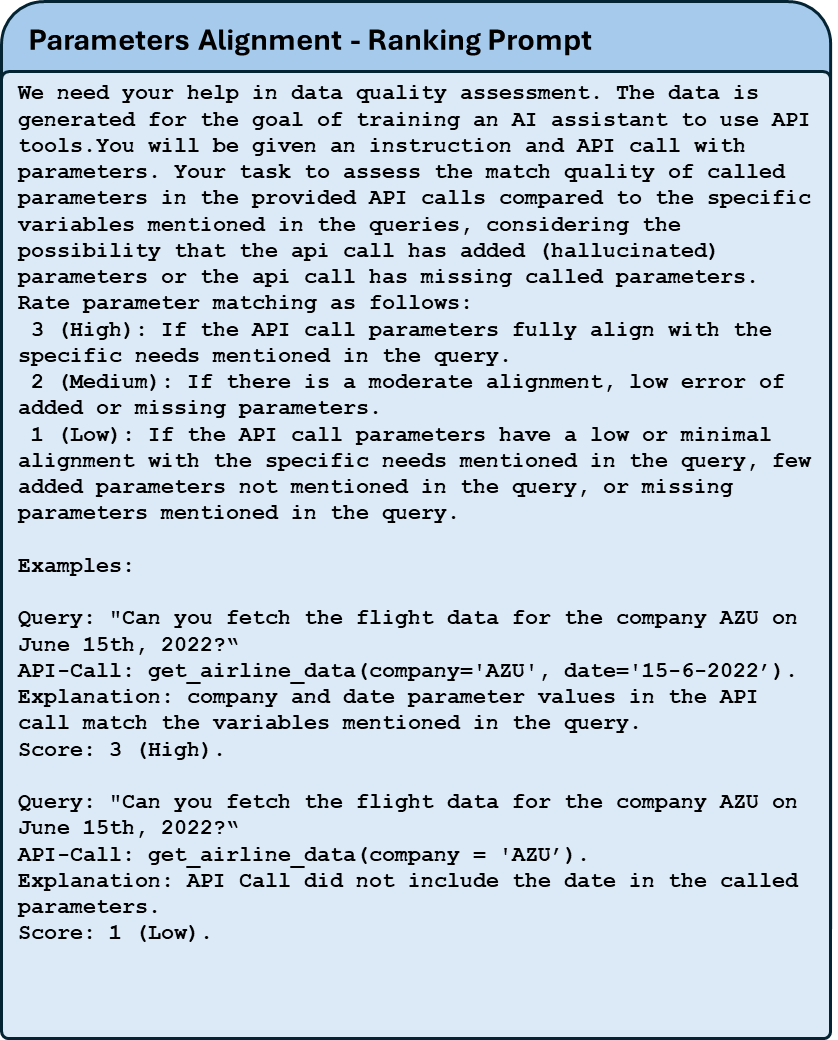}
  \caption{Prompt for \textbf{parameter alignment}, as a direct questioning task.
  }
  \label{fig:prompt_param_direct}
\end{figure}

Direct questioning and annotation instruction with \evaluatorModel{} did not work well for the criteria of \textit{Specificity}, \textit{Coherence} and \textit{Parameter Alignment}. In Figures \ref{fig:prompt_specificity_direct}--\ref{fig:prompt_param_direct} we provide the prompts. In \autoref{tab:failed_direct} we provide validation results of alignment with human annotation on the same subset of examples of the ToolBench dataset. 

\begin{table}[htbp]
    \centering
    \begin{tabular}{ccccc}
    \toprule
    \textbf{Criterion} & \textbf{Acc.} & \textbf{Precision} & \textbf{Recall} & \textbf{F1}  \\
    \midrule
     Specificity & 0.54 & 0.56 & 0.36 & 0.43  \\
     Coherence & 0.74 & 0.50 & 0.46 & 0.48 \\
     Alignment & 0.66 & 0.69 & 0.71 & 0.70 \\
    \bottomrule
    \end{tabular}
    \caption{Validation results when using the direct questioning approach for  \textit{Specificity}, \textit{Coherence} and \textit{Parameter Alignment}. Compare to \autoref{tab:intrinsic_metrics}, which shows higher scores for the prompts ultimately used.}
    \label{tab:failed_direct}
\end{table}


\subsection{Manual Annotation Process}
\label{appendix_manual_annotation}

\subsubsection{Annotating Training Data}
\label{sec:annotation_train}
To initiate the annotation process, we examined the data and identified the quality criteria (as outlined in \S\ref{sec_intrinsic_eval_framework}). We then went through several cycles of examination and refinement of respective guidelines. 

An instance of annotation shows the instruction, the available API functions, and the API-call sequence that should solve the instruction. The annotator needs to mark level of specificity of the instruction (1 to 3), its coherence (1 to 3), whether it is solvable with respect to the available API functions (yes/no), the sequence call validity in terms of function availability (yes/no), parameter alignment in the calls (yes/no), whether the sequence call solves the instruction (yes/no), and whether it does so minimally (yes/no). See Tables \ref{tab:annotation_guidelines_specificity}, \ref{tab:annotation_guidelines_coherence} and \ref{tab:annotation_guidelines_solvability} for annotation instructions of the first three criteria.

The annotators (authors of this paper) first annotated the same 20 instances from ToolBench and discussed differences, culminating in strong agreement between the annotators.
The averaged Kappa statistics for the first three criteria are: Specificity 0.674 (``substantial''), Coherence 0.508 (``moderate''), and Solvability 0.414 (``moderate'').
Annotators were then assigned different samples of data, for a total of 50 instances from the ToolBench train set, and 50 from the ToolAlpaca train set. We used this data to assess the intrinsic metrics that we developed (\S\ref{sec_intrinsic_eval_metrics}).



\subsubsection{Annotating the ToolBench Test Set}
In comparison to annotation of training instances, the test set annotation differs in two major aspects. First, the test set does not include API-call sequences, but rather only the input instructions. A training instance consists of an API-call sequence in order to teach an LLM how to devise a solution for attaining a final result. However during test time, tool-assisted LLMs are typically evaluated on the final result, and not on the API-call sequence used to achieve the result.
Second, in our cleaned test set, we do not only mark inadequate instances, but we also attempt to fix instructions so that they become usable. The ToolBench test set contains 1100 instances (distinct from the 125K instances), and only filtering out faulty instances would leave very few suitable ones. Essentially, we use the ToolBench test set as data to build upon instead of creating new data altogether, which would be a much costlier procedure. The ultimate goal is to produce a high-quality test set of solvable multi-request instructions.

An instruction can fail on either specificity, coherence or solvability. Therefore, to repair an instruction we focused on the failing criteria and rewrote the instruction to mend the faults. We allowed for some creativity as long as the quality criteria were intact, and the same number of requests was kept within the instruction.

For example, 
\textit{``I'm planning a family movie night and I want to watch some classic films. Can you suggest some iconic movies available on YouTube? Also, find a YouTube playlist of movie soundtracks. Additionally, provide the latest versions of C++, Objective-C, and Scala programming languages for my cousin who is a software developer.''} Here, the first request (``suggest iconic movies'') and the second request (``find a YouTube playlist'') are not specific enough for the available API functions, and the third request (``provide the latest versions of C++...'') is not coherent with the beginning of the instruction. We therefore rewrote the instruction for this instance as \textit{``I'm learning how to program and I'd like some assistance. Can you suggest some videos on YouTube about C++? Also, download the video to MP3 from `www.youtube.com/?123abc'. Additionally, please let me know the the latest versions of C++, Objective-C, and Scala programming languages.''} The new instruction resolves the three issues described. In a case where it is unclear how to use the respective available API functions, no fix is made and the instance is simply discarded.

Five annotators annotated 674 of the 1100 instances in the ToolBench test set. $27.6\%$ of the instances lacked specificity, $21.5\%$ lacked coherence, and $32.7\%$ were unsolvable. Overall, $37.7\%$ of the instances were discarded, in cases where errors were too severe to be readily fixable.
The new test set is used for measuring the performance of tool-using models in the multi-request setting (\S\ref{sec_extrinsic_eval}), and can generally be used as a high-quality benchmark. We provide the new test set in the supplementary material.

\begin{table}[h!]
    \centering
    \begin{tabular}{|p{0.95\columnwidth}|}
        \hline
        \textbf{Specificity} \\
        Evaluate the extent to which the data examples contain all necessary information without gaps or missing variables for the AI assistant to address the user requests. \\ \hline
        \textbf{1 (Poor):} The instruction is extremely broad and general, lacking essential information. \\ \hline
        \textbf{2 (Medium):} The instruction includes moderate specific details but there are some gaps in information. \\ \hline
        \textbf{3 (Excellent):} The instruction is highly specific and complete, with no significant missing information. \\ \hline
    \end{tabular}
    \caption{Human annotation guidelines for Specificity.}
    \label{tab:annotation_guidelines_specificity}
\end{table}

\begin{table}[h!]
    \centering
    \begin{tabular}{|p{0.95\columnwidth}|}
        \hline
        \textbf{Coherence} \\
        Evaluate the extent to which the different requests in the instruction are logically connected and relevant to each other. \\ \hline
        \textbf{1 (Poor):} The different requests of the instruction are highly disjointed, lacking a logical connection. \\ \hline
        \textbf{2 (Medium):} The different requests of the instruction have a moderate level of coherence but still possess some degree of separation. \\ \hline
        \textbf{3 (Excellent):} The components of the instruction are highly coherent, with a strong logical connection. \\ \hline
        \textbf{Not Applicable:} When there is only one request. (Considered as ‘3’ for filtering.) \\ \hline
    \end{tabular}
    \caption{Human annotation guidelines for Coherence.}
    \label{tab:annotation_guidelines_coherence}
\end{table}

\begin{table}[h!]
    \centering
    \begin{tabular}{|p{0.95\columnwidth}|}
        \hline
        \textbf{Solvability} \\
        Determine if the ground truth APIs can handle the instruction in terms of functionality. It is alright if a parameter value is not explicitly provided in the query. \\ \hline
        \textbf{0 (No):} The request cannot be handled by the given APIs. The APIs' functionalities do not fit or address the request. \\ \hline
        \textbf{1 (Yes):} The instruction can be handled by using the given APIs. A parameter value might not be explicitly provided in the query. \\ \hline
    \end{tabular}
    \caption{Human annotation guidelines for Solvability.}
    \label{tab:annotation_guidelines_solvability}
\end{table}

\subsection{Qualitative Examples}
\label{app:qualitative_examples}
In Tables \ref{tab:qualitative_examples_specificity} and \ref{tab:qualitative_examples_coherence} we provide examples of instructions which our method found as lacking specificity and coherence, from both ToolBench and ToolAlpaca datasets.
\begin{table*}[ht]
\centering
\renewcommand{\arraystretch}{1.5} 
\begin{tabular}{|c|p{0.9\linewidth}|}
\hline
\multicolumn{2}{|l|}{\textbf{Instruction Examples}} \\
\hline
\multirow{5}{*}{\rotatebox[origin=c]{90}{\textbf{From ToolBench}}} & I'm planning to buy a used car and I need to decode the VIN number of a specific vehicle. Can you provide me with the car model, maker, year, engine, and other relevant information? Additionally, I'm curious about the trending search results on Google. \\
\cline{2-2}
& I'm a wedding planner and I want to create personalized videos for my clients. Can you give me the details of a specific template I have in mind, including the variables it offers? Also, I need to access all my campaigns' information, including the images, videos, and image+video campaigns. \\
\cline{2-2}
& I'm hosting a garden party next weekend. Can you give me the 1-hour/minutely forecast for the party location? Additionally, recommend some outdoor games and decorations for the event. \\
\cline{2-2}
& I recently discovered a new song that I really love. Can you provide me with the lyrics and related data for the song? Also, suggest some similar songs that I might enjoy. \\
\cline{2-2}
& I'm planning a trip to Europe and I want to stay updated on the energy prices in the region. Can you fetch all the available articles from a specific region, like Europe? Additionally, provide me with a list of news sources and their corresponding regions. \\
\hline
\multirow{5}{*}{\rotatebox[origin=c]{90}{\textbf{From ToolAlpaca}}} & Please generate an invoice for my freelance work and send it to my client. \\
\cline{2-2}
& How can I find the best gear for my character in Guild Wars 2? \\
\cline{2-2}
& Hey, I'm planning a road trip and I want to check for any road closures along my route. Can you help me with that? \\
\cline{2-2}
& I need to retrieve detailed information about a specific malware sample. Can you show me how to do that? \\
\cline{2-2}
& I want to know if any of the email addresses in a list are disposable. Can you use the API to check which email addresses in the list are disposable? \\

\hline
\end{tabular}
\caption{Examples of instructions which our method found as lacking \textbf{specificity}, from the two examined datasets.}
\label{tab:qualitative_examples_specificity}
\end{table*}


\begin{table*}[ht]
\centering
\begin{tabular}{|c|p{0.9\linewidth}|}
\hline
\multicolumn{2}{|l|}{\textbf{Instruction Examples}} \\
\hline
\multirow{6}{*}{\rotatebox[origin=c]{90}{\textbf{From ToolBench}}} & I'm planning a surprise birthday party for my best friend and I need some help. Can you find the email of a person named Emma Watson at google.com? Additionally, I want to find a formulated product by its registration number to use as a gift for my friend. \\
\cline{2-2}
& My family and I are considering relocating to New York City. Can you provide us with a list of transactions for zipcode 10019? We would like to see the last sales date, last sales amount, and total records for each transaction. Additionally, could you give us the detailed historical transactions for the address 310 W 56th St, New York, NY 10019? \\
\cline{2-2}
& I want to explore movies related to a specific genre. Can you discover movies in the genre with genreId '80' and provide me with the details of the first 10 results? Also, fetch the crew details for a random movie. \\
\cline{2-2}
& I'm a basketball enthusiast and I want to know more about the players in the NBA. Can you fetch me the details of all the players? Additionally, provide me with a random Chuck Norris joke to lighten the mood. \\
\cline{2-2}
& My friends and I are planning a trip to multiple cities and we need to estimate the cost of living. Can you provide us with a list of available currencies? Additionally, we would like to get a comprehensive list of cities, including their countries, to help us plan our itinerary. \\
\hline
\multirow{6}{*}{\rotatebox[origin=c]{90}{\textbf{From ToolAlpaca}}} & I'm curious about quotes related to debugging. Can you find some for me? After that, please show me a list of all authors so I can learn more about their thoughts on programming. \\
\cline{2-2}
& I want to add a catchy animation to my GitHub profile. Show me a list of font types available for use, and once I choose one, create a typing and deleting SVG with the text "I'm a software engineer" in 18-point font size, orange color, a typing speed of 80 ms, start delay of 500 ms, and a pause duration of 1 second. \\
\cline{2-2}
& I'm thinking of going to Lansdowne Park this afternoon. Could you find nearby bus stops within a 300-meter radius with my current location at latitude 45.3967 and longitude -75.6858? \\
\cline{2-2}
& My user profile still shows my old email address. Can you update it to my new one, "new\_email@example.com"? Also, update my preferences to receive newsletters about datasets in the "economy" category. \\
\cline{2-2}
& Can you personalize the email content for my subscribers based on their names? Use the template 'Holiday Greetings' and add subscriber data for Sarah, whose email is sarah@example.com and name is 'Sarah Smith'. \\
\hline
\end{tabular}
\caption{Examples of instructions which our method found as \textbf{incoherent}, from the two examined datasets.}
\label{tab:qualitative_examples_coherence}
\end{table*}

\subsection{Relationship Between Quality Criteria}
\label{app:dimensions_relations}
\begin{figure}[h]
    \centering
    \begin{subfigure}[b]{0.45\textwidth}
        \centering
        \includegraphics[width=\textwidth]{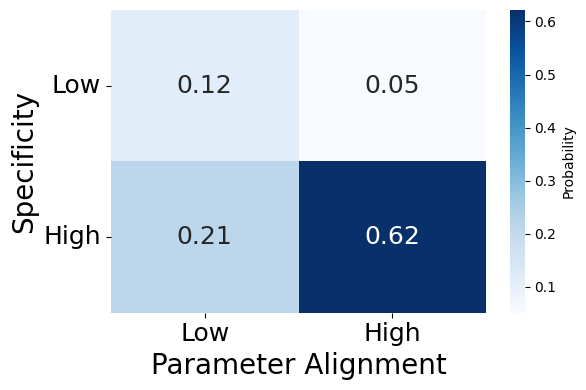}
        \caption{ToolAlpaca dataset}
        \label{fig:alpaca_confusion_matrix}
    \end{subfigure}
    \hspace{0.05\textwidth}
    \begin{subfigure}[b]{0.45\textwidth}
        \centering
        \includegraphics[width=\textwidth]{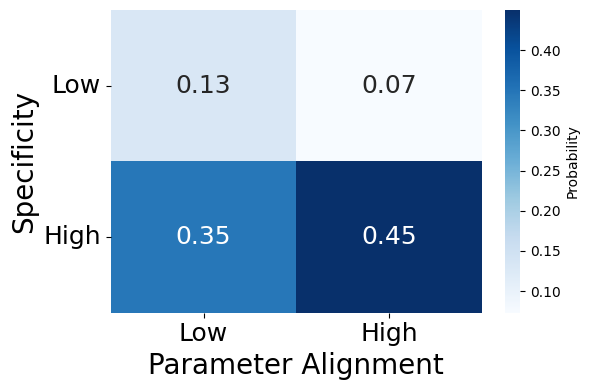}
        \caption{ToolBench dataset}
        \label{fig:bench_confusion_matrix}
    \end{subfigure}
    \caption{Confusion matrices for Specificity and Parameter Alignment.}
    \label{fig:spec_param_conf}
\end{figure}

We additionally explored the relationship between quality criteria within the datasets. Generally, the correlations between dimensions are not particularly high. A notable analysis we conducted shows the effect of \textit{Specificity} on \textit{Parameter Alignment}. As illustrated in \autoref{fig:spec_param_conf}, when specificity is weak, it is also more likely that parameter alignment is weak. This might be expected behavior since low specificity means that parameter values are missing in the instruction, and the LLM hallucinates a value in order to complete its task. The correlation however is not exceedingly high, in particular we see in ToolBench that even for instances with high specificity, the parameter alignment can still be low, showing that there are examples where the parameter is present in the instruction but it does not match the parameter in the ground-truth response.

\section{ICE}
\label{app:ice}

\subsection{Full Prompt}
In \autoref{fig:ice_prompt} we provide the full prompt for our proposed in-context evaluation method. The prompt is constructed as follows: a description of the task, documentation of the APIs selected, one in-context example and the test queries.

\subsection{ICE Score Calculation}
\begin{figure*}[h]
  \centering
  \includegraphics[width=0.95\textwidth,height=1.4\textwidth]{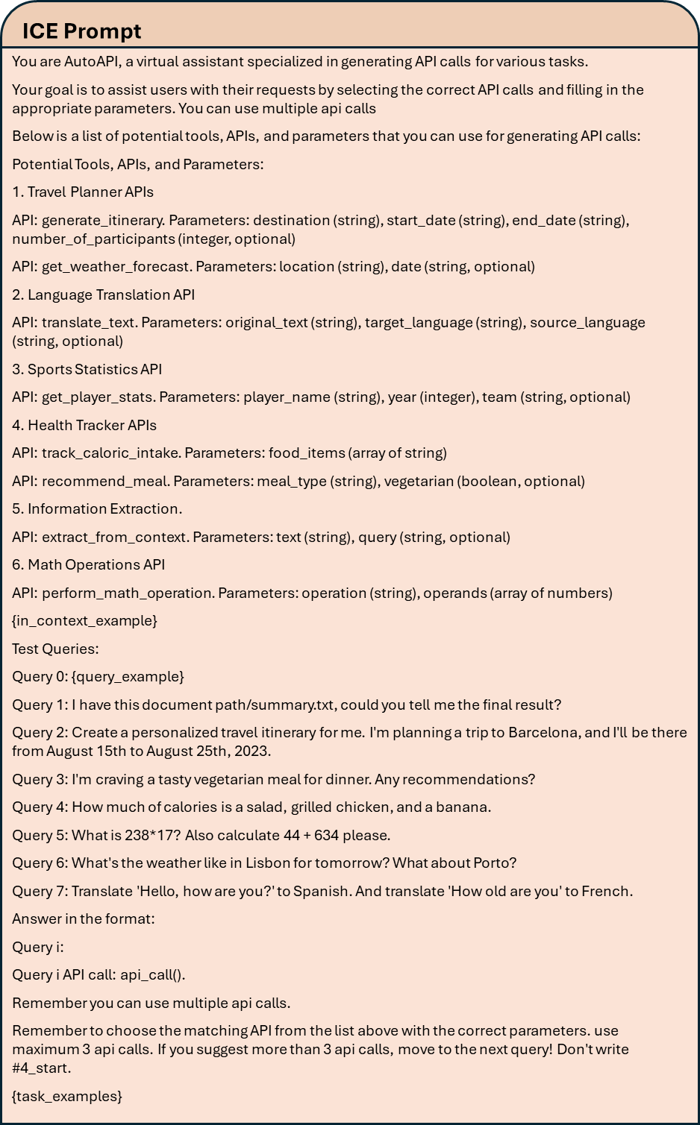}
  \caption{The prompt used for in-context evaluation of a training instance (marked as \texttt{\{in\_context\_example\}} in the prompt).}
  \label{fig:ice_prompt}
\end{figure*}

To calculate the ICE score, we follow these steps:

\begin{enumerate}
    \item We input to the model the ICE prompt (\autoref{fig:ice_prompt}), containing an in-context example from the assessed dataset, and obtain the model output for each of the 7 test instructions.
    \item For each test instruction, we calculate the Levenshtein distance between the generated API-call sequence and the correct API-call sequence.
    \item We average the Levenshtein distances calculated for all test instructions, resulting in a single score for each data instance.
\end{enumerate}

Steps 1 to 3 are repeated for each of the data instances in the assessed dataset. \autoref{fig:ice_distribution} shows the distribution of instance-level scores for the two assessed datasets.

 \section{Extrinsic Evaluation}
 \label{app:extrinsic}
 
 \subsection{Training Setup}
 \label{app:training}
The training setup is similar for both ToolBench and ToolAlpaca benchmarks, where we train on pairs of (instruction, API-call sequence + response).
 
\paragraph{ToolBench.}
We fine-tune a LLaMA-7B model when working with the ToolBench dataset. The learning rate is set to $5 \times 10^{-5}$, and we use a batch size of 2. Since the tasks require relatively long inputs for the targeted model, the context length is extended using positional interpolation \cite{chen2023extending}. We increase the context length to 4096, which is twice the model's default length of 2048. The model is trained for two epochs on 8 NVIDIA A10G Tensor Core GPUs.
\paragraph{ToolAlpaca.}
For the ToolAlpaca dataset, we fine-tune a Vicuna-7B model. We use a batch size of 2 and a learning rate of $2 \times 10^{-5}$. The model is fine-tuned for three epochs on 4 NVIDIA A10G Tensor Core GPUs.

\subsection{Evaluation Setup}
\label{app:evaluation}

We adhere to the evaluation procedures outlined in the respective benchmarks for ToolBench and ToolAlpaca. Both benchmarks use a generative model for the evaluation of the API-call sequence and response. We use ChatGPT for both datasets.

\paragraph{ToolBench.}
In the ToolBench benchmark, the evaluation process begins with assessing the solvability of the given instruction.  Using ChatGPT, solution paths are categorized as Pass, Fail, or Unsure based on this classification. The evaluation criteria include various rules to determine the success of a solution path. For more detailed insights into the evaluation methodology and rules, please refer to the original paper \cite{qin2024toolllm}.

The original evaluation procedure involves assessing the generalization ability across three levels---unseen instructions, tools, and categories---as well as three different scenarios. However, instead of splitting the test set into categories, we calculate the pass rate by averaging over all test samples. Importantly, in the human-annotation of the test set, we aimed to maintain a similar distribution across all test splits for consistency.

Regarding the retrieval of APIs during model inference, we adopt only one of the approaches tested in the original evaluation, where we directly insert the relevant APIs for each test instruction. This approach simulates the scenario where the user specifies the preferred API set.

\paragraph{ToolAlpaca.}
Similarly, in the ToolAlpaca benchmark, we use ChatGPT to evaluate the model's output in addressing the instruction. The evaluation criteria is assessing the overall correctness, considered as the pass rate, of both the process and the response. For further details regarding the evaluation methodology, please refer to the original paper \cite{tang2023toolalpaca}.

In our study, we use the simulated subset for evaluation. This subset comprises 10 simulated tools (100 instructions) that were not part of the training toolset. While the original paper also includes a real-world subset with 11 APIs from various domains, we focused solely on the simulated data due the lack of detailed instructions on how to use the real-world data.

 \newpage


\newpage

\clearpage

\end{document}